\documentclass[11pt,letterpaper]{article}
\usepackage{emnlp2016}
\usepackage{times}
\usepackage{latexsym}

\PassOptionsToPackage{numbers,compress}{natbib}

\usepackage{color}

\usepackage{graphicx}

\usepackage{subcaption}

\usepackage[olditem,oldenum]{paralist}

\usepackage{bbm}

\usepackage{multirow}
\usepackage{rotating}
\usepackage{booktabs}

\newcommand{\figref}[1]{\figurename~\ref{#1}}

\usepackage[pagebackref=true,breaklinks=true,bookmarks=false,colorlinks=true,citecolor=blue]{hyperref}

\emnlpfinalcopy

\def\emnlppaperid{629}

\newcommand{\ad}{\textcolor{black}}

\title{Human Attention in Visual Question Answering: \\
Do Humans and Deep Networks Look at the Same Regions?}

\author{
    Abhishek Das$^1$\thanks{\,\, Denotes equal contribution.}\,\,,
    Harsh Agrawal$^1$\footnotemark[1]\,\,,
    C. Lawrence Zitnick$^2$,
    Devi Parikh$^1$,
    Dhruv Batra$^1$\\
    $^1$Virginia Tech, Blacksburg, VA 24061, USA\\
    {\tt\small\{abhshkdz, harsh92, parikh, dbatra\}@vt.edu}\\
    $^2$Facebook AI Research, Menlo Park, CA 94025, USA\\
    {\tt\small zitnick@fb.com}
}

\date{}

\begin{document}

\maketitle

\begin{abstract}
  We conduct large-scale studies on `human attention' in Visual Question Answering (VQA) to understand where humans choose to look to answer questions about images.
We design and test multiple game-inspired novel attention-annotation interfaces that require the subject to sharpen regions of a blurred image to answer a question.
Thus, we introduce the VQA-HAT (Human ATtention) dataset.
We evaluate attention maps generated by state-of-the-art VQA models against human attention both qualitatively (via visualizations) and quantitatively (via rank-order correlation).
Overall, our experiments show that current attention models in VQA do not seem to be looking at the same regions as humans.
\end{abstract}

\section{Introduction}
\label{sec:intro}

It helps to pay attention.
Humans have the ability to quickly perceive a scene by selectively attending to parts of the image instead of processing the whole scene in its entirety \cite{rensink_vc2000}.
Inspired by human attention, a recent trend in computer vision and deep learning is to build computational models of attention.
Given an input signal, these models learn to attend to parts of it for further processing and have been successfully applied in machine translation \cite{bahdanau_iclr15,firat_arxiv16}, object recognition \cite{ba_iclr15,mnih_arxiv14,sermanet_arxiv14}, image captioning \cite{xu_arxiv15,cho_arxiv15} and visual question answering \cite{yang_arxiv15,lu_arxiv16,hxu_arxiv15,xiong_arxiv16}.

\begin{figure}[t]
    \includegraphics[width=1\linewidth]{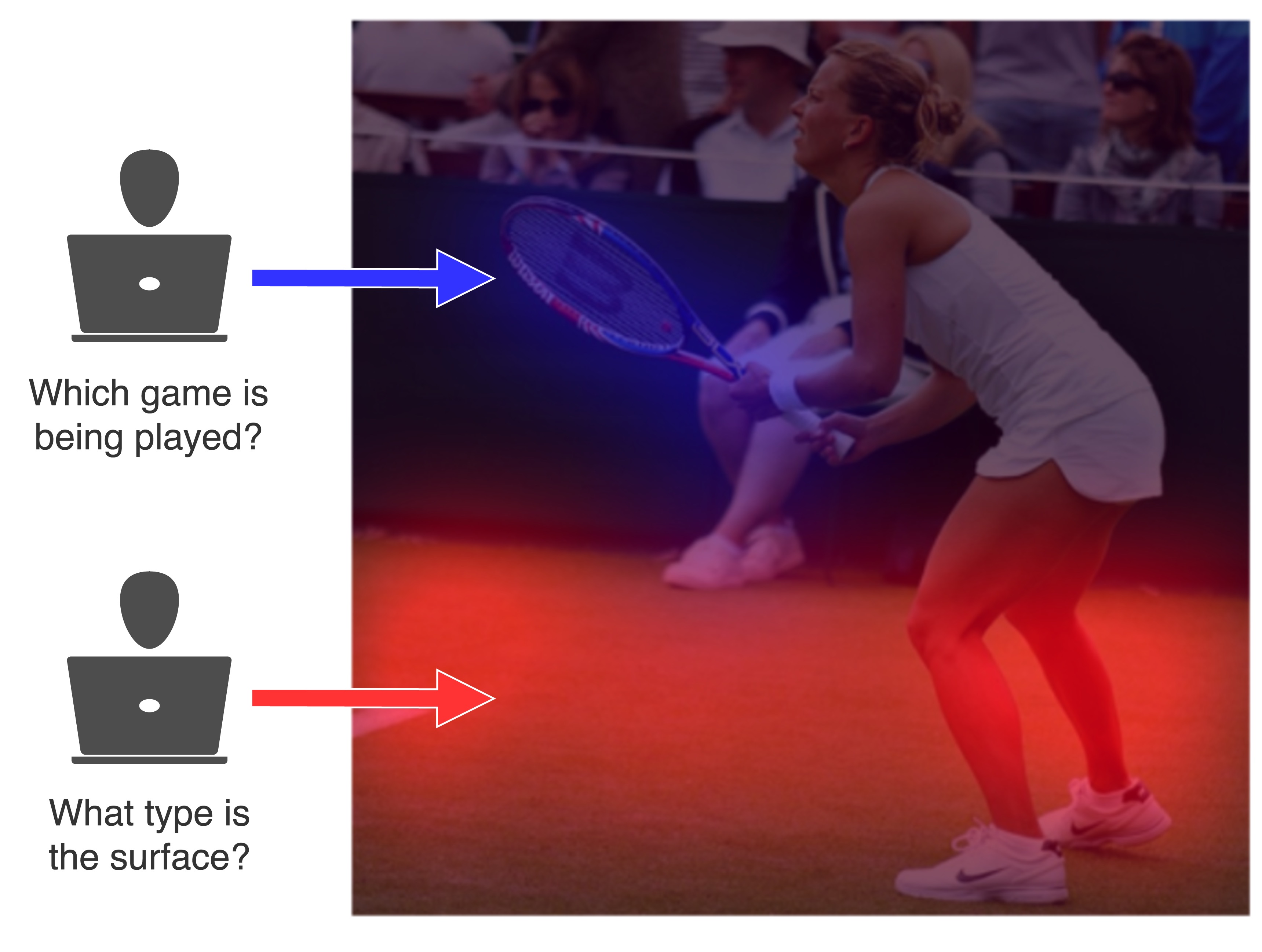}
    \caption{Different human attention regions based on question (best viewed in color).}
    \label{fig:teaser}
\end{figure}

In this work, we study attention for the task of Visual Question Answering (VQA).
Unlike image captioning, where a coarse understanding of an image is often sufficient for producing generic descriptions \cite{devlin_arxiv15}, visual questions selectively target different areas of an image including background details and underlying context.
This suggests that a VQA model may benefit from an explicit or implicit attention mechanism to answer a question correctly.

\begin{figure*}[ht]
  \centering
  \begin{subfigure}[b]{0.32\textwidth}
    \includegraphics[width=1\linewidth]{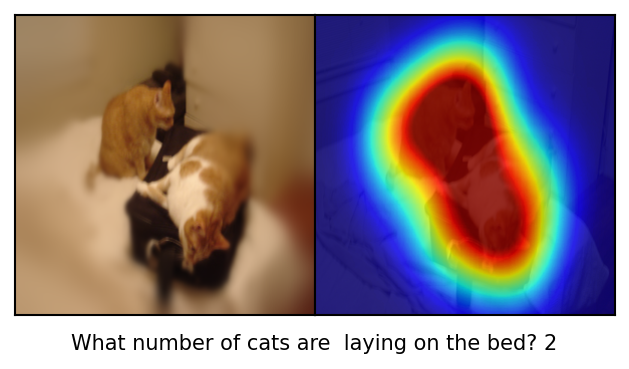}
    \caption{}
  \end{subfigure}
  \begin{subfigure}[b]{0.32\textwidth}
    \includegraphics[width=1\linewidth]{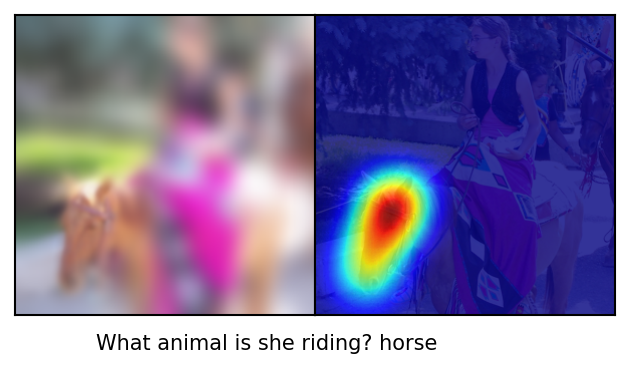}
    \caption{}
  \end{subfigure}
  \begin{subfigure}[b]{0.32\textwidth}
    \includegraphics[width=1\linewidth]{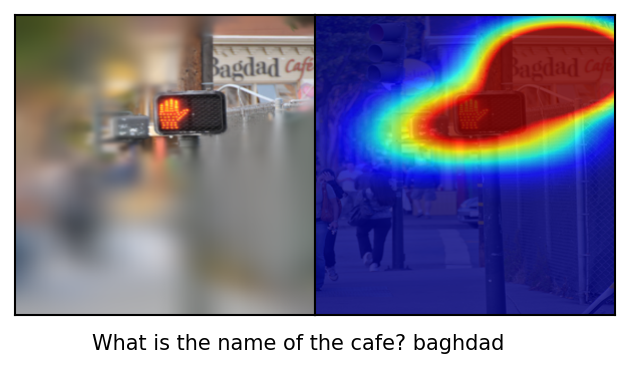}
    \caption{}
  \end{subfigure}
  \caption{
    (a-c): Column 1 shows deblurred image, and column 2 shows human attention map.}
  \label{fig:human_maps}
\end{figure*}


In this work, we are interested in the following questions:
1) Which image regions do humans choose to look at in order to answer questions about images?
2) Do deep VQA models with attention mechanisms attend to the same regions as humans?

We design and conduct studies to collect ``human attention maps".
\figref{fig:teaser} shows human attention maps on the same image for two different questions.
When asked `What type is the surface?', humans choose to look at the floor, while attention for `Which game is being played?' is concentrated around the player and racket.


These human attention maps can be used both for evaluating machine-generated attention maps and for explicitly training attention-based models.

\noindent \textbf{Contributions.}
First, we design and test multiple game-inspired novel interfaces for collecting human attention maps of where humans choose to look to answer questions from the large-scale VQA dataset \cite{antol_iccv15}; this VQA-HAT (Human ATtention) dataset will be released publicly.
Second, we perform qualitative and quantitative comparison of the maps generated by state-of-the-art attention-based VQA models \cite{yang_arxiv15,lu_arxiv16} and a task-independent saliency baseline \cite{judd_iccv09} against our human attention maps through visualizations and rank-order correlation.
We find that machine-generated attention maps from the most accurate VQA model have a mean rank-correlation of 0.26 with human attention maps, which is worse than task-independent saliency maps that have a mean rank-correlation of 0.49.
It is well understood that task-independent saliency maps have a `center bias'~\cite{tatler_jov07,judd_iccv09}.
After we control for this center bias in our human attention maps, we find that the correlation of task-independent saliency is poor (as expected), while trends for machine-generated VQA-attention maps remain the same (which is promising).

\section{Related Work}
\label{sec:related}





\begin{figure*}[h!t]
    \centering
    \begin{subfigure}[b]{0.32\textwidth}
        \includegraphics[width=1\textwidth]{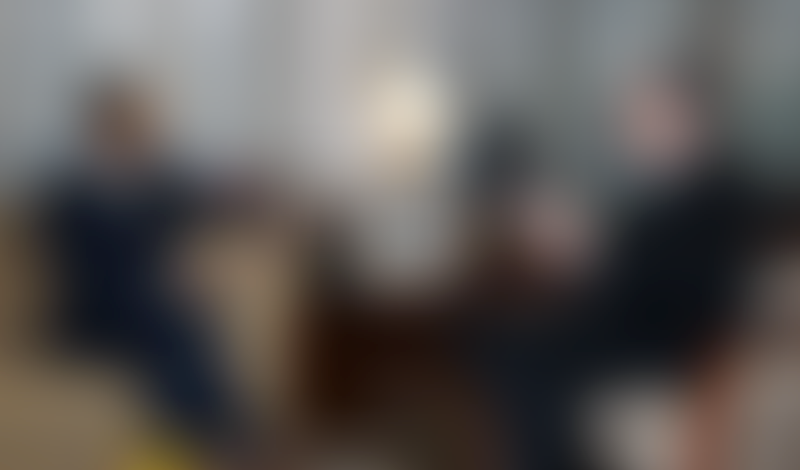}%
        \caption{Initial blurred image}
        \label{fig:step1}
    \end{subfigure}
    \begin{subfigure}[b]{0.32\textwidth}%
        \includegraphics[width=1\textwidth]{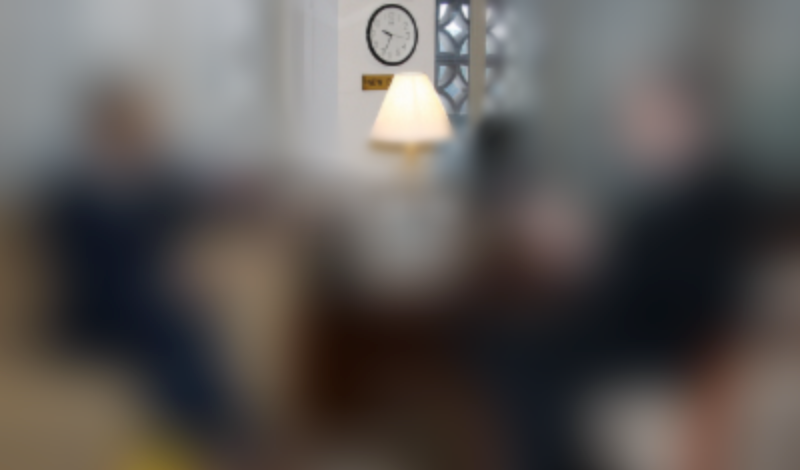}
        \caption{Regions sharpened by subject}
        \label{fig:step2}
    \end{subfigure}
    \begin{subfigure}[b]{0.32\textwidth}
        \includegraphics[width=1\textwidth]{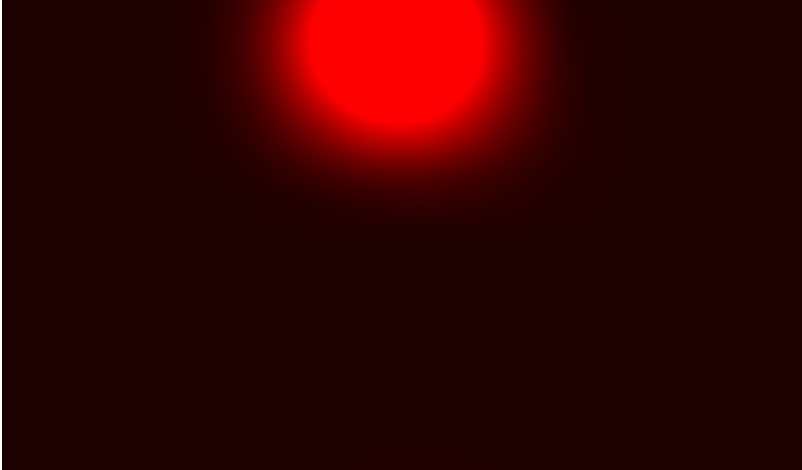}%
        \caption{Attention map}
        \label{fig:step3}
    \end{subfigure}
    \caption{Deblurring procedure to collect attention maps.
    \ad{We present subjects with a blurred image and ask them to sharpen regions of the image that will help them answer the question correctly, in a smooth, click-and-drag, `coloring' motion with the mouse.}} \vspace{-10pt}
\label{fig:task_steps}
\end{figure*}

Our work draws on recent work in attention-based VQA and human studies in saliency prediction.
We work with the free-form and open-ended VQA dataset released by \cite{antol_iccv15}.


\noindent \textbf{VQA Models.} Attention-based models for VQA typically use convolutional neural networks to highlight relevant regions of image given a question.
Stacked Attention Networks (SAN) proposed in \cite{yang_arxiv15} use LSTM encodings of question words to produce a spatial attention distribution over the convolutional layer features of the image.
Hierarchical Co-Attention Network \cite{lu_arxiv16} generates multiple levels of image attention based on words, phrases and complete questions, and is the top entry on the VQA Challenge\footnote{\url{http://visualqa.org/challenge.html}} as of the time of this submission.
Another interesting approach uses question parsing to compose the neural network from modules, attention being one of the sub-tasks addressed by these modules \cite{andreas_arxiv16}.

Note that all these works are \emph{unsupervised} attention models, where ``attention'' is simply an intermediate variable (a spatial distribution) that is produced by the model to optimize downstream loss (VQA cross-entropy).
The fact that some (it's unclear how many) of these spatial distributions end up being interpretable is simply fortuitous.
In contrast, we study where humans choose to look to answer visual questions.
These human attention maps can be used to evaluate unsupervised maps.

\noindent \textbf{Human Studies.} There's a rich history of work in collecting eye tracking data from human subjects to gain an understanding of image saliency and visual perception \cite{jiang_eccv14,judd_iccv09,feifei_jov07,yarbus_1967}.
Eye tracking data to study natural visual exploration \cite{jiang_eccv14,judd_iccv09} is useful but difficult and expensive to collect on a large scale.
\cite{jiang_cvpr15} established mouse tracking as an accurate approach to collecting attention maps.
They collected large-scale attention annotations for MS COCO \cite{coco} on Amazon Mechanical Turk (AMT).
While \cite{jiang_cvpr15} studies natural exploration and collects task-independent human annotations by asking subjects to freely move the mouse cursor to anywhere they wanted to look on a blurred image, our approach is task-driven.


Specifically, as described in \ref{sec:dataset}, we collect ground truth attention annotations by instructing subjects to sharpen parts of a blurred image that are important for answering the questions accurately.
Section \ref{sec:experiments} covers evaluation of unsupervised attention maps generated by VQA models against our human attention maps.
\section{VQA-HAT (Human ATtention) Dataset}
\label{sec:dataset}

\begin{figure*}[hp]
    \centering
    \begin{subfigure}[b]{0.8\textwidth}
        \includegraphics[width=1\textwidth]{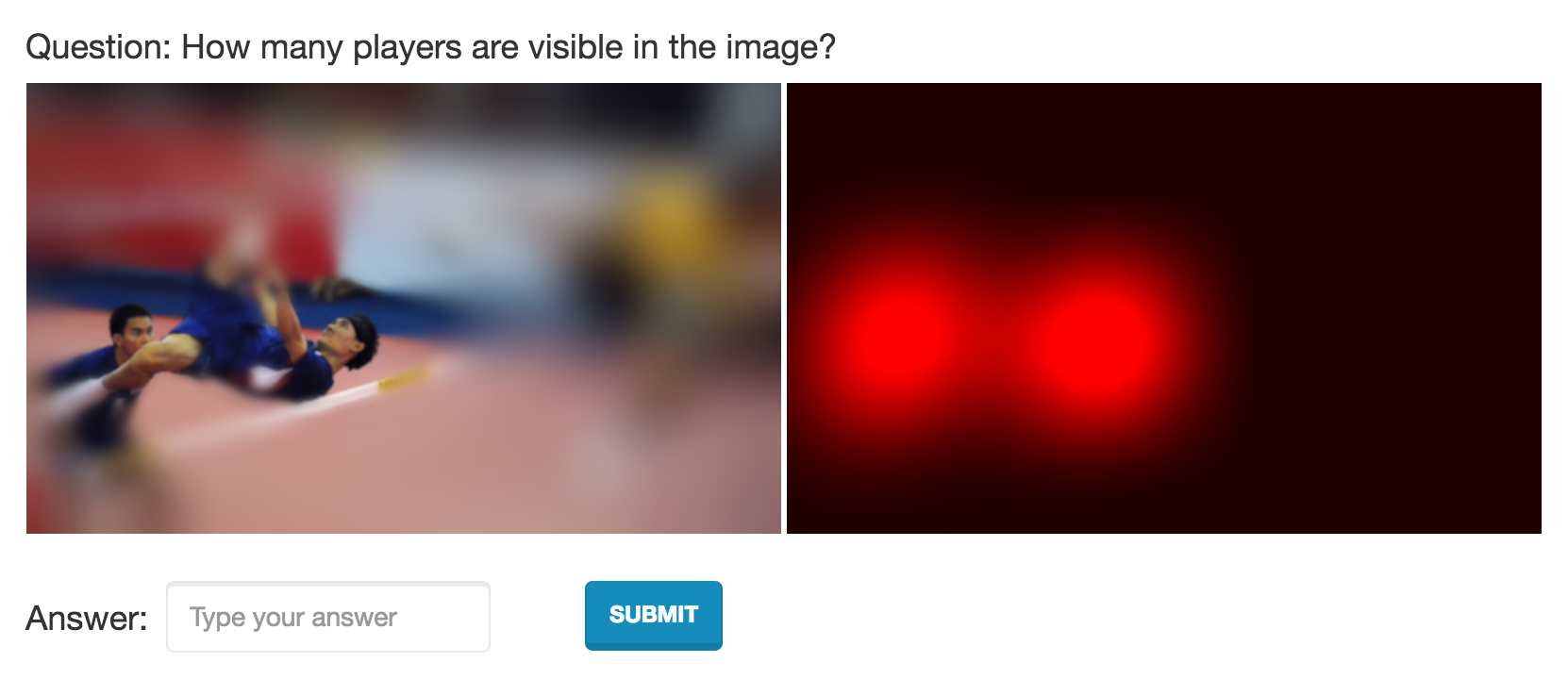}
        \caption{Blurred Image without Answer}
        \label{fig:i1}
    \end{subfigure}

    \begin{subfigure}[b]{0.8\textwidth}
        \includegraphics[width=1\textwidth]{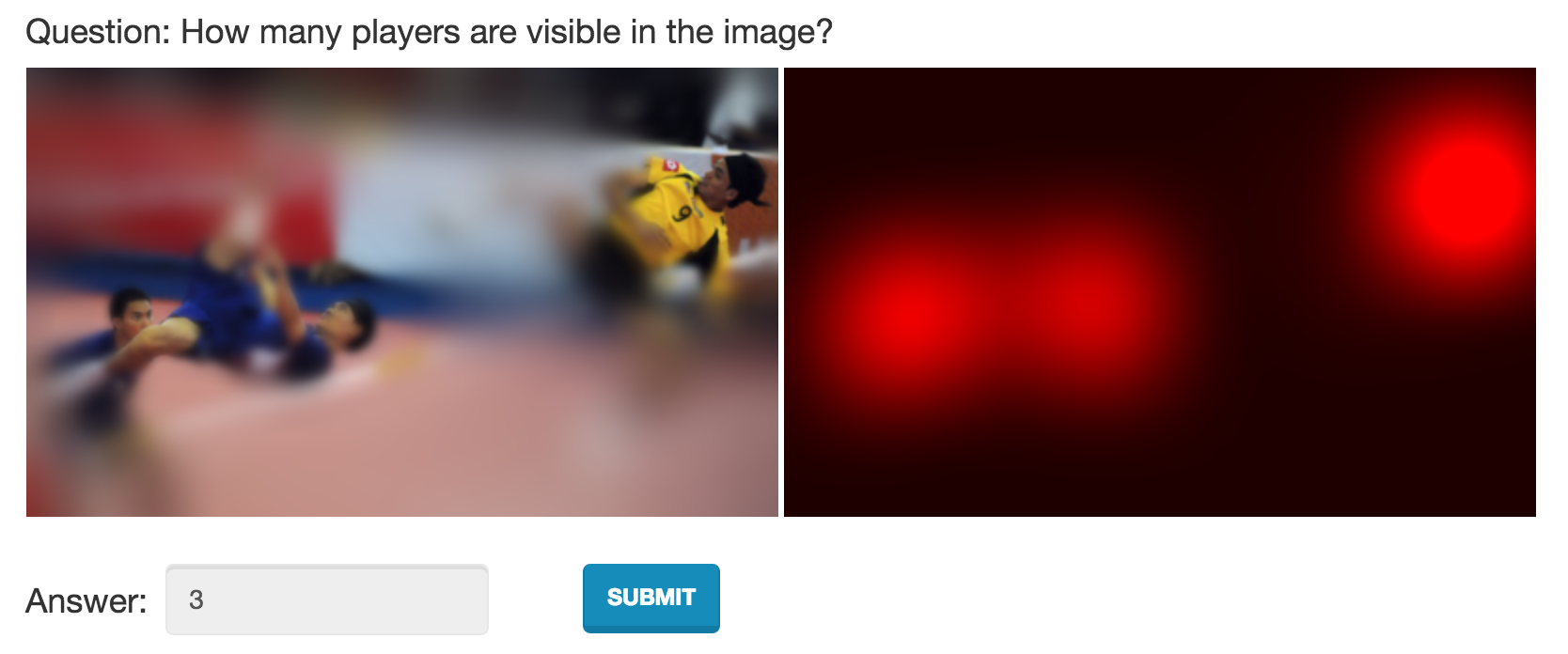}
        \caption{Blurred Image with Answer}
        \label{fig:i2}
    \end{subfigure}

    \begin{subfigure}[b]{0.8\textwidth}
        \includegraphics[width=1\textwidth]{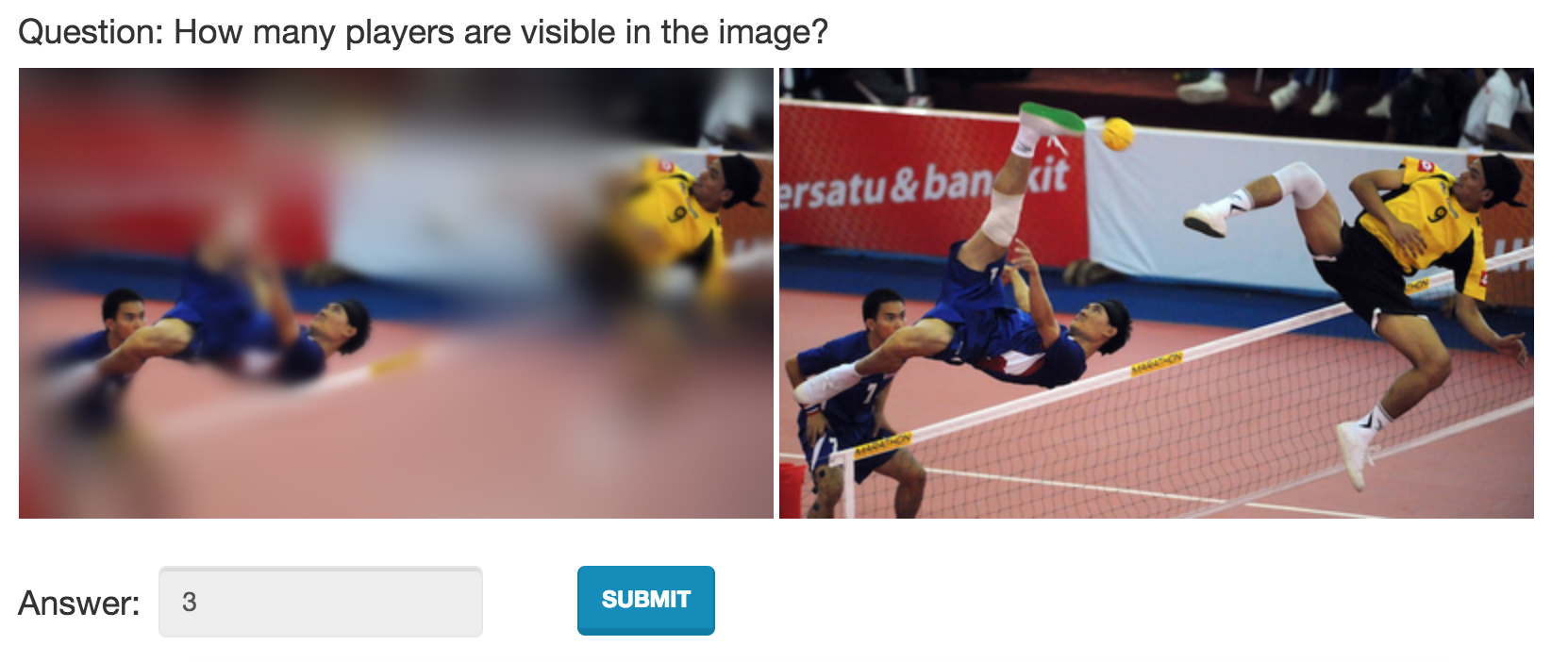}
        \caption{Blurred \& Original Image with Answer}
        \label{fig:i3}
    \end{subfigure}
    \caption{Attention annotation interface variants.
    (a) In our first interface, subjects were shown a blurred image and a question without the answer, and were asked to deblur regions and enter the answer.
    (b) In our second interface, subjects were shown the correct answer in addition to the question and blurred image.
They were asked to sharpen as few regions as possible such that someone can answer the question just by looking at the blurred image with sharpened regions.
    (c) To encourage exploitation instead of exploration, in our third interface, subjects were shown the question-answer pair and full-resolution original image.
    Out of the three interfaces, “Blurred Image with Answer” (b) struck the right balance between exploration and exploitation, and gives the highest accuracy on evaluation by humans as described in section \ref{sec:evaluation}.} \vspace{-10pt}
\label{fig:interfaces}

\end{figure*}

We design and test multiple game-inspired novel interfaces for conducting large-scale human studies on AMT.
Our basic interface design consists of a ``deblurring" exercise for answering visual questions.
Specifically, we present subjects with a blurred image and a question about the image, and ask subjects to sharpen regions of the image that will help them answer the question correctly, in a smooth, click-and-drag, `coloring' motion with the mouse.
The sharpening is gradual: successively scrubbing the same region progressively sharpens it.
\figref{fig:task_steps} shows intermediate steps in our attention annotation interface, from a completely blurry image to a deblurred attention map.

\subsection{Attention Annotation Interface} \label{sec:interface}

Our interface starts by showing a low-resolution blurry version of the image.
This is to convey a partial `holistic' understanding of the scene to the subjects so they may intelligently choose which regions to sharpen.
Gradual sharpening with strokes was aimed to capture initial exploration as they tried to get a better sense of the scene, and eventually focussed sharpening to answer the question.
Next we describe the three variants of our attention annotation interface that we experimented with.

\subsubsection{Blurred Image without Answer} \label{sec:interface_1}

In our first interface, subjects were shown a blurred image and a question without the answer, and were asked to deblur regions and enter the answer.
We found that this interface sometimes resulted in `exploratory attention', where the subject lightly sharpens large regions of an image to find salient regions that eventually lead them to the answer.
However, subjects often ended up with `incomplete' attention maps since they did not see the high-resolution image and the answer, so they did not know when to stop deblurring or exploring.
For instance, for an image with 3 players playing a sport, if the question is ``How many players are visible in the image?", the subject might sharpen a region that seems to have the players, count the 2 players in there and answer 2, and completely miss another region of the image that had 1 more.
The resulting attention map in this case is incomplete since there are 3 players in the image.
This effect of incomplete human attention maps was seen in counting (``How many ...") and binary (``Is there ...") types of questions, and as a result, the answers to these were often incorrect.

\subsubsection{Blurred Image with Answer} \label{sec:interface_2}

In our second interface, subjects were shown the correct answer in addition to the question and blurred image.
They were asked to sharpen as few regions as possible such that someone can answer the question just by looking at the blurred image with sharpened regions.
This interface is shown in \figref{fig:i2}.
Providing the answer fixed the failure cases from the 1st interface, i.e.\ for counting and binary questions, since the subjects now knew the answer, they continued to explore till they found the answer region in the image.

\subsubsection{Blurred and Original Image with Answer} \label{sec:interface_3}

To encourage exploitation instead of exploration, in our third interface, subjects were shown the question-answer pair and full-resolution original image.
In principle, seeing the original (full-resolution) image, the question, and answer provides most information to subjects, thus enabling them to provide the most `accurate' attention maps.
However, this task turns out to be fairly counter-intuitive -- subjects are shown full-resolution images and the answer, and asked to imagine a scenario where someone else has to answer the question without looking at the original image.

\figref{fig:interfaces} shows screen-captures of the 3 attention annotation interfaces.

\subsection{Dataset Evaluation} \label{sec:evaluation}

We ran pilot studies on AMT to experiment with the above described three interfaces.
In order to quantitatively evaluate the interfaces, we conducted a second human study where (a second set of) subjects where shown the attention-sharpened images generated from each of the attention interfaces from the first experiment and asked to answer the question.
The intuition behind this experiment is that if the attention map revealed too little information, this second set of subjects would answer the question incorrectly.
Table~\ref{tab:human_acc} shows VQA accuracies of the answers given by human subjects under these 3 interfaces.
We can see that the ``Blurred Image with Answer" interface (section \ref{sec:interface_2}) gives the highest accuracy on evaluation by humans.

Since the payments structure on AMT encourage completing tasks as quickly as possible, this implicitly incentivizes subjects to deblur as few regions as possible, and our human study shows that humans can still answer questions.
Thus, overall we achieve a balance between highlighting too little or too much.


\begin{table}[h]
\centering
\resizebox{0.98\columnwidth}{!}{%
\begin{tabular}{ccc}
\toprule
Interface Type                          & Human Accuracy \\ \midrule
Blurred Image without Answer            & 75.2        \\
Blurred Image with Answer               & 78.7        \\
Blurred \& Original Image with Answer   & 71.2         \\
Original Image                          & 80.0         \\ \bottomrule
\end{tabular}%
}
\caption{Human accuracies to compare the quality of human attention maps collected by different interfaces.
Subjects were shown deblurred images from each of these interfaces and asked to answer the visual question.}
\label{tab:human_acc}
\end{table}

We collected human attention maps for 58475 train (out of 248349 total) and 1374 val (out of 121512 total) question-image pairs in the VQA dataset.
Overall, we conducted approximately 20000 Human Intelligence Tasks (HITs) on AMT, among 800 unique workers.
\figref{fig:human_maps} shows examples of collected human attention maps.
This VQA-HAT dataset will be released publicly.

\begin{figure}[h!tp]
    \includegraphics[width=1\linewidth]{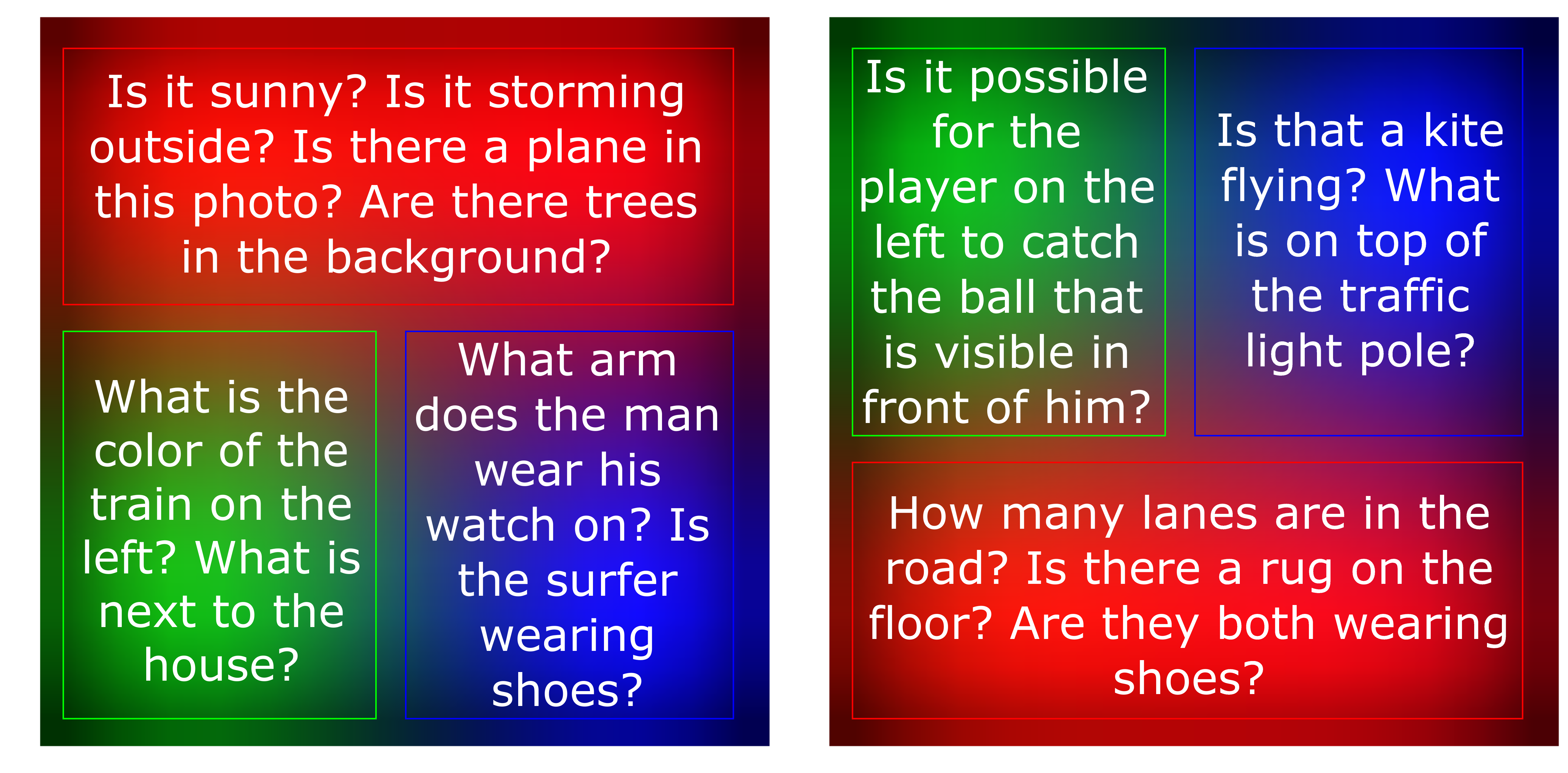}
    \caption{Visualization of 6 human attention map clusters -- the average attention map for the cluster and example questions falling in each of them.}
    \label{fig:analysis}
\end{figure}

To visualize the collected dataset, we cluster the human attention maps and visualize the average attention map and example questions falling in each of them for 6 selected clusters in \figref{fig:analysis}.

\section{Human Attention Maps vs Unsupervised Attention Models}
\label{sec:experiments}

\begin{figure*}[hp]
    \includegraphics[width=1\linewidth]{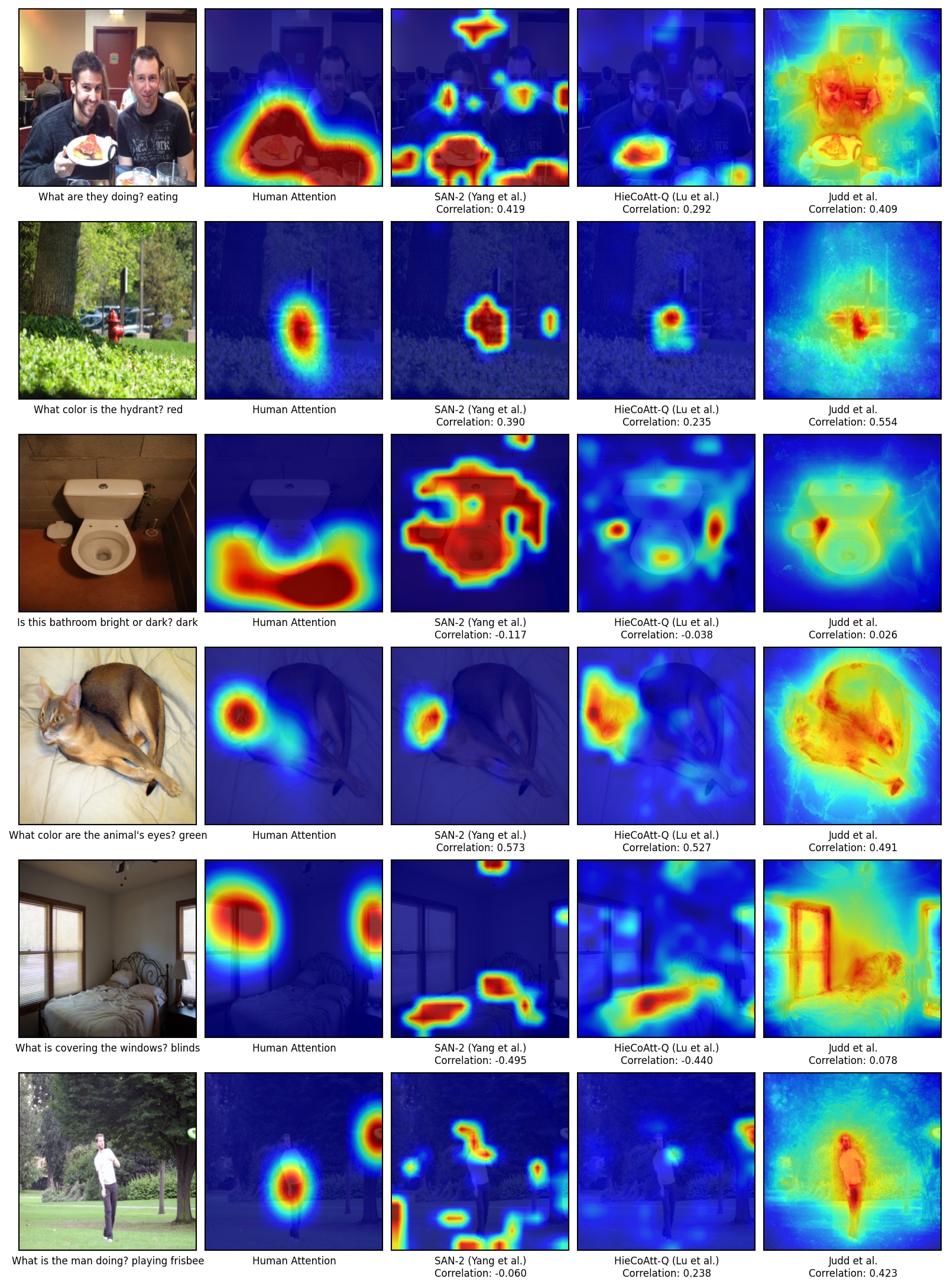}
    \caption{Random samples of human attention (column 2) v/s machine-generated attention (columns 3-5).}
    \label{fig:eval}
\end{figure*}

Now that we have collected these human attention maps, we can ask the following question -- do unsupervised attention models learn to predict attention maps that are similar to human attention maps?
To rephrase, \emph{do neural networks look at the same regions as humans to answer a visual question?}

\noindent \textbf{VQA Attention Models.} We evaluate maps generated by the following unsupervised models:
\begin{compactitem}
    \item Stacked Attention Network (SAN) \cite{yang_arxiv15} with two attention layers (SAN-2)\footnote{Code available at \url{https://github.com/zcyang/imageqa-san}.}.
    \item Hierarchical Co-Attention Network (HieCoAtt) \cite{lu_arxiv16} with word-level (HieCoAtt-W), phrase-level (HieCoAtt-P) and question-level (HieCoAtt-Q) attention maps; we evaluate all three maps\footnote{Code available at \url{https://github.com/jiasenlu/HieCoAttenVQA}}.
\end{compactitem}


\noindent \textbf{Comparison Metric: Rank Correlation.} We first scale both the machine-generated and human attention maps to 14x14, rank the pixels according to their spatial attention and then compute correlation between these two ranked lists.
We choose an order-based metric so as to make the evaluation invariant to absolute spatial probability values which can be made peaky or diffuse by tweaking a `temperature' parameter.

\begin{table}[h]
\resizebox{0.98\columnwidth}{!}{%
\begin{tabular}{ccc}
\toprule
Model & Rank-correlation \\ \midrule
SAN-2 \cite{yang_arxiv15}  & 0.249 $\pm$ 0.004 \\ \cmidrule(lr){1-2}
HieCoAtt-W \cite{lu_arxiv16} & 0.246 $\pm$ 0.004 \\
HieCoAtt-P \cite{lu_arxiv16} & 0.256 $\pm$ 0.004 \\
HieCoAtt-Q \cite{lu_arxiv16} & 0.264 $\pm$ 0.004 \\ \cmidrule(lr){1-2}
Random & 0.000 $\pm$ 0.001 \\ \cmidrule(lr){1-2}
Judd et al. \cite{judd_iccv09} & 0.497 $\pm$ 0.004 \\ \cmidrule(lr){1-2}
Human & 0.623 $\pm$ 0.003 \\ \bottomrule
\end{tabular}
}
\caption{Mean rank-correlation coefficients (higher is better); error bars show standard error of means.
We can see that both SAN-2 and HieCoAtt attention maps are positively correlated with human attention maps, but not as strongly as task-independent Judd saliency maps.}
\label{tab:eval}
\end{table}

\noindent Table~\ref{tab:eval} shows rank-order correlation averaged over all image-question pairs on the validation set.
We compare with random attention maps and task-independent saliency maps generated by a model trained to predict human eye fixation locations where subjects are asked to freely view an image for 3 seconds \cite{judd_iccv09}.
Both SAN-2 and HieCoAtt attention maps are positively correlated with human attention maps, but not as strongly as task-independent Judd saliency maps.
Our findings lead to two take-away messages with significant potential impact on future research in this active field.
First, current VQA attention models do not seem to be `looking' at the same regions as humans to produce an answer.
\ad{Second, as attention-based VQA models become more accurate ($58.9\%$ SAN $\rightarrow$ $62.1\%$ HieCoAtt), they seem to be (slightly) better correlated with humans in terms of where they look.}
Our dataset will allow for a more thorough validation of this observation as future attention-based VQA models are proposed.
\figref{fig:eval} shows examples of human attention and machine-generated attention maps with corresponding rank-correlation coefficients.

To put these numbers in perspective, we computed inter-human agreement on the validation set by collecting 3 human attention maps per image-question pair and computing mean rank-correlation, which is 0.623.
Lastly, all reported correlation values are averaged over 3 trials by adding random noise (order of $10^{-14}$) to the human attention maps to account for ranking variations in case of uniformly weighted regions.



\noindent \textbf{Center Bias.} Judd saliency maps aim to predict human eye fixations during natural visual exploration.
These tend to have a strong center bias \cite{tatler_jov07,judd_iccv09}.
Although our human attention maps dataset is not an eye tracking study, the center bias still exists albeit not as severe.
One potential source of this center bias is the fact that the VQA dataset was human-generated by subjects looking at the images.
Thus, salient objects in the center of the image are likely be potential subjects of the questions.
We compute rank-correlation of a synthetically generated central attention map with Judd saliency and human attention maps.
Judd saliency maps have a mean rank-correlation of 0.877 and human attention maps have a mean rank-correlation of 0.458 on the validation set.


\begin{table}[h!]
\resizebox{0.98\columnwidth}{!}{%
\begin{tabular}{ccc}
\toprule
Model & Rank-correlation \\ \midrule
SAN-2 \cite{yang_arxiv15}  & 0.038 $\pm$ 0.011 \\ \cmidrule(lr){1-2}
HieCoAtt-W \cite{lu_arxiv16} & 0.062 $\pm$ 0.012 \\
HieCoAtt-P \cite{lu_arxiv16} & 0.048 $\pm$ 0.010 \\
HieCoAtt-Q \cite{lu_arxiv16} & 0.114 $\pm$ 0.012 \\ \cmidrule(lr){1-2}
Judd et al. \cite{judd_iccv09} & -0.063 $\pm$ 0.009 \\ \bottomrule
\end{tabular}
}
\caption{Mean rank-correlation coefficients (higher is better) on the reduced set without center bias; error bars show standard error of means.
We can see that correlation goes down significantly for Judd saliency maps since they have a strong center bias.
Relative trends among SAN-2 \& HieCoAtt are similar to those over the whole validation set (reported in Table~\ref{tab:eval}).
}
\label{tab:eval_centerless}
\end{table}

To eliminate the effect of center bias in this evaluation, we removed human attention maps that have a positive rank-correlation with the center attention map.
We compute rank-correlation of machine-generated attention with human attention on this reduced set.
See Table~\ref{tab:eval_centerless}.
Mean correlation goes down significantly for Judd saliency maps since they have a strong center bias.
Relative trends among SAN-2 \& HieCoAtt are similar to those over the whole validation set (reported in Table~\ref{tab:eval}).
HieCoAtt-Q now has a higher correlation with human attention maps than Judd saliency.
This demonstrates that discounting the center bias, VQA-specific machine attention maps correlate better with VQA-specific human attention maps than task independent machine saliency maps.


\section{Conclusion \& Discussion}
\label{sec:discussion}



We introduce and release the VQA-HAT dataset.
This dataset can be used to evaluate attention maps generated in an unsupervised manner by attention-based VQA models, or to explicitly train models with attention supervision for VQA.
We quantify whether current attention-based VQA models are `looking' at the same regions of the image as humans do to produce an answer.

\noindent \textbf{Necessary vs Sufficient Maps.} Are human attention maps `necessary' and/or `sufficient'?
If regions highlighted by the human attention maps are sufficient to answer the question accurately, then so is any region that is a superset.
For example, if attention mass is concentrated on a `cat' for `What animal is present in the picture?', then an attention map that assigns weights to any arbitrary-sized region that includes the `cat' is sufficient as well.
On the contrary, a \emph{necessary} and sufficient attention map would be the smallest visual region sufficient for answering the question accurately.
It is an ill-posed problem to define a necessary attention map in the space of pixels; random pixels can be blacked out and chances are that humans would still be able to answer the question given the resulting subset attention map.
Our work thus poses an interesting question for future work -- what is the right \emph{semantic} space in which it is meaningful to talk about necessary and sufficient attention maps for humans?
\section{Acknowledgements}

We thank Jiasen Lu and Ramakrishna Vedantam for helpful suggestions and discussions.
This work was supported in part by the following:
National Science Foundation CAREER awards to DB and DP,
Army Research Office YIP awards to DB and DP,
ICTAS Junior Faculty awards to DB and DP,
Army Research Lab grant W911NF-15-2-0080 to DP and DB,
Office of Naval Research grant N00014-14-1-0679 to DB,
Paul G. Allen Family Foundation Allen Distinguished Investigator award to DP,
Google Faculty Research award to DP and DB,
AWS in Education Research grant to DB, and NVIDIA GPU donation to DB.

{\small
\bibliography{main}

\begin{thebibliography}{}

\bibitem[\protect\citename{Andreas \bgroup et al.\egroup
  }2016]{andreas_arxiv16}
Jacob Andreas, Marcus Rohrbach, Trevor Darrell, and Dan Klein.
\newblock 2016.
\newblock Learning to compose neural networks for question answering.
\newblock {\em CoRR}, abs/1601.01705.

\bibitem[\protect\citename{Antol \bgroup et al.\egroup }2015]{antol_iccv15}
Stanislaw Antol, Aishwarya Agrawal, Jiasen Lu, Margaret Mitchell, Dhruv Batra,
  C.~Lawrence Zitnick, and Devi Parikh.
\newblock 2015.
\newblock Vqa: Visual question answering.

\bibitem[\protect\citename{Ba \bgroup et al.\egroup }2015]{ba_iclr15}
Jimmy~Lei Ba, Volodymyr Mnih, and Koray Kavukcuoglu.
\newblock 2015.
\newblock {Multiple Object Recognition With Visual Attention}.
\newblock {\em Iclr-2015}.

\bibitem[\protect\citename{Bahdanau \bgroup et al.\egroup
  }2014]{bahdanau_iclr15}
Dzmitry Bahdanau, Kyunghyun Cho, and Yoshua Bengio.
\newblock 2014.
\newblock Neural machine translation by jointly learning to align and
  translate.
\newblock {\em CoRR}, abs/1409.0473.

\bibitem[\protect\citename{Cho \bgroup et al.\egroup }2015]{cho_arxiv15}
KyungHyun Cho, Aaron~C. Courville, and Yoshua Bengio.
\newblock 2015.
\newblock Describing multimedia content using attention-based encoder-decoder
  networks.
\newblock {\em CoRR}, abs/1507.01053.

\bibitem[\protect\citename{Devlin \bgroup et al.\egroup }2015]{devlin_arxiv15}
Jacob Devlin, Saurabh Gupta, Ross Girshick, Margaret Mitchell, and C.~Lawrence
  Zitnick.
\newblock 2015.
\newblock {Exploring Nearest Neighbor Approaches for Image Captioning}.
\newblock {\em arXiv preprint}.

\bibitem[\protect\citename{Fei-Fei \bgroup et al.\egroup }2007]{feifei_jov07}
Li~Fei-Fei, Asha Iyer, Christof Koch, and Pietro Perona.
\newblock 2007.
\newblock What do we perceive in a glance of a real-world scene?
\newblock {\em Journal of Vision}, 7(1):10.

\bibitem[\protect\citename{Firat \bgroup et al.\egroup }2016]{firat_arxiv16}
Orhan Firat, KyungHyun Cho, and Yoshua Bengio.
\newblock 2016.
\newblock Multi-way, multilingual neural machine translation with a shared
  attention mechanism.
\newblock {\em CoRR}, abs/1601.01073.

\bibitem[\protect\citename{Jiang \bgroup et al.\egroup }2014]{jiang_eccv14}
Ming Jiang, Juan Xu, and Qi~Zhao.
\newblock 2014.
\newblock {Saliency in Crowd}.
\newblock {\em ECCV}.

\bibitem[\protect\citename{Jiang \bgroup et al.\egroup }2015]{jiang_cvpr15}
Ming Jiang, Shengsheng Huang, Juanyong Duan, and Qi~Zhao.
\newblock 2015.
\newblock Salicon: Saliency in context.
\newblock In {\em CVPR}, June.

\bibitem[\protect\citename{Judd \bgroup et al.\egroup }2009]{judd_iccv09}
Tilke Judd, Krista Ehinger, Fr{\'e}do Durand, and Antonio Torralba.
\newblock 2009.
\newblock Learning to predict where humans look.
\newblock In {\em ICCV}.

\bibitem[\protect\citename{Lin \bgroup et al.\egroup }2014]{coco}
Tsung-Yi Lin, Michael Maire, Serge Belongie, James Hays, Pietro Perona, Deva
  Ramanan, Piotr Dollár, and C.~Lawrence Zitnick.
\newblock 2014.
\newblock Microsoft {COCO}: Common objects in context.

\bibitem[\protect\citename{{Lu} \bgroup et al.\egroup }2016]{lu_arxiv16}
J.~{Lu}, J.~{Yang}, D.~{Batra}, and D.~{Parikh}.
\newblock 2016.
\newblock {Hierarchical Co-Attention for Visual Question Answering}.
\newblock {\em ArXiv e-prints}, May.

\bibitem[\protect\citename{Mnih \bgroup et al.\egroup }2014]{mnih_arxiv14}
Volodymyr Mnih, Nicolas Heess, Alex Graves, and Koray Kavukcuoglu.
\newblock 2014.
\newblock {Recurrent Models of Visual Attention}.
\newblock {\em arXiv preprint}.

\bibitem[\protect\citename{Rensink}2000]{rensink_vc2000}
Ronald~A. Rensink.
\newblock 2000.
\newblock The dynamic representation of scenes.
\newblock {\em Visual Cognition}, 7(1-3):17--42.

\bibitem[\protect\citename{Sermanet \bgroup et al.\egroup
  }2014]{sermanet_arxiv14}
Pierre Sermanet, Andrea Frome, and Esteban Real.
\newblock 2014.
\newblock Attention for fine-grained categorization.
\newblock {\em CoRR}, abs/1412.7054.

\bibitem[\protect\citename{Tatler}2007]{tatler_jov07}
Benjamin~W. Tatler.
\newblock 2007.
\newblock The central fixation bias in scene viewing: Selecting an optimal
  viewing position independently of motor biases and image feature
  distributions.
\newblock {\em Journal of Vision}, 7(14):4.

\bibitem[\protect\citename{Xiong \bgroup et al.\egroup }2016]{xiong_arxiv16}
Caiming Xiong, Stephen Merity, and Richard Socher.
\newblock 2016.
\newblock Dynamic memory networks for visual and textual question answering.
\newblock {\em CoRR}, abs/1603.01417.

\bibitem[\protect\citename{Xu and Saenko}2015]{hxu_arxiv15}
Huijuan Xu and Kate Saenko.
\newblock 2015.
\newblock Ask, attend and answer: Exploring question-guided spatial attention
  for visual question answering.
\newblock {\em CoRR}, abs/1511.05234.

\bibitem[\protect\citename{Xu \bgroup et al.\egroup }2015]{xu_arxiv15}
Kelvin Xu, Jimmy Ba, Ryan Kiros, Kyunghyun Cho, Aaron~C. Courville, Ruslan
  Salakhutdinov, Richard~S. Zemel, and Yoshua Bengio.
\newblock 2015.
\newblock Show, attend and tell: Neural image caption generation with visual
  attention.
\newblock {\em CoRR}, abs/1502.03044.

\bibitem[\protect\citename{Yang \bgroup et al.\egroup }2015]{yang_arxiv15}
Zichao Yang, Xiaodong He, Jianfeng Gao, Li~Deng, and Alexander~J. Smola.
\newblock 2015.
\newblock Stacked attention networks for image question answering.
\newblock {\em CoRR}, abs/1511.02274.

\bibitem[\protect\citename{Yarbus}1967]{yarbus_1967}
A.~L. Yarbus.
\newblock 1967.
\newblock {\em Eye Movements and Vision}.
\newblock Plenum. New York.

\end{thebibliography}
\bibliographystyle{emnlp2016}
}

\end{document}